\documentclass[letterpaper, 10 pt, conference]{ieeeconf}

\IEEEoverridecommandlockouts                              

\usepackage{amsmath} 
\usepackage{amssymb}  
\usepackage{multirow}
\usepackage{array}
\usepackage{amsmath}
\usepackage{gensymb}
\usepackage{graphicx}
\usepackage{hyperref}
\usepackage[]{subcaption}
\usepackage{amsmath}
\usepackage{amsfonts}
\usepackage[linesnumbered,algoruled,boxed]{algorithm2e}
\usepackage{mathtools}
\usepackage[normalem]{ulem}
\usepackage{cancel}
\usepackage{subfloat}
\usepackage{lipsum}

\usepackage{cite}
\usepackage{hyperref}
\hypersetup{colorlinks,linktocpage=true}

\newcommand{\bX}{\mathbf{X}}

\newcommand{\br}{\mathbf{r}}
\newcommand{\bt}{\mathbf{t}}
\newcommand{\bx}{\mathbf{x}}
\newcommand{\btheta}{\boldsymbol{\theta}}
\newcommand{\bphi}{\boldsymbol{\phi}}
\newcommand{\bPi}{\boldsymbol{\Pi}}

\newcommand{\rb}{\mathrm{b}}
\newcommand{\rc}{\mathrm{c}}
\newcommand{\rw}{\mathrm{w}}

\newcommand{\csT}[3]{\prescript{#1}{#2}{#3}}

\newcommand{\cscam}[1]{\prescript{\rc}{}{#1}}
\newcommand{\csw}[1]{\prescript{\rw}{}{#1}}

\newcolumntype{P}[1]{>{\centering\arraybackslash}p{#1}}
\title{
ROVO: Robust Omnidirectional Visual Odometry \\for Wide-baseline Wide-FOV Camera Systems
}

\author{Hochang Seok and Jongwoo Lim$^*$\\
{\tt\small \{hochangseok, jlim\}@hanyang.ac.kr}\\
{\small Department of Computer Science, Hanyang University, Seoul, Korea.}
}

\begin{document}

\maketitle
\thispagestyle{empty}
\pagestyle{empty}

\begin{abstract}

In this paper we propose a robust visual odometry system for a wide-baseline camera rig with wide field-of-view (FOV) fisheye lenses, which provides full omnidirectional stereo observations of the environment. 
For more robust and accurate ego-motion estimation we adds three components to the standard VO pipeline, 1) the hybrid projection model for improved feature matching, 2) multi-view P3P RANSAC algorithm for pose estimation, and 3) online update of rig extrinsic parameters. 
The hybrid projection model combines the perspective and cylindrical projection to maximize the overlap between views and minimize the image distortion that degrades feature matching performance. 
The multi-view P3P RANSAC algorithm extends the conventional P3P RANSAC to multi-view images so that all feature matches in all views are considered in the inlier counting for robust pose estimation. 
Finally the online extrinsic calibration is seamlessly integrated in the backend optimization framework so that the changes in camera poses due to shocks or vibrations can be corrected automatically.
The proposed system is extensively evaluated with synthetic datasets with ground-truth and real sequences of highly dynamic environment, and its superior performance is demonstrated.

\end{abstract}

 
\section{INTRODUCTION}

Ego-motion estimation is an essential functionality in achieving autonomous navigation and maneuver of robots.
To estimate the robot's pose in the environment the surrounding structures and objects needs to be modeled accurately. 
Recently many approaches using various sensors including LIDARs, radars, and/or cameras have been developed.
Among them the cameras have advantages for their low cost, passive sensing, abundant information on the environment, mechanical robustness, and many more. 

Visual odometry (VO), ego-motion estimation using one or more cameras, has been researched for a few decades to accomplish real-time processing, accurate pose estimation, and robustness to the external disturbances.
It has been widely applied to many applications including augmented/virtual reality (AR/VR), advanced driver assistance system (ADAS) and autonomous driving.
There exist variety of VO systems, for example ORB-SLAM2~\cite{mur2017orb} is monocular or stereo and using feature points with binary descriptors, or Stereo-DSO~\cite{wang2017stereo} uses a stereo camera and edges for the feature. 
Monocular systems are attractive for their simple hardware configuration, but it has several limitations that the true scale of motion cannot be estimated, and it is mostly use for AR/VR where metric poses are not needed.
Multi-view (stereo) VO systems produce more robust and true metric ego-motion estimates suitable for robot applications, but requires more computational cost.
All VO algorithms are vulnerable to excessive dynamic objects, i.e., when the significant portion of the field-of-view (FOV) is covered by moving objects, the pose estimation becomes unstable and incorrect.

\begin{figure}[t]
\includegraphics[width=\linewidth]{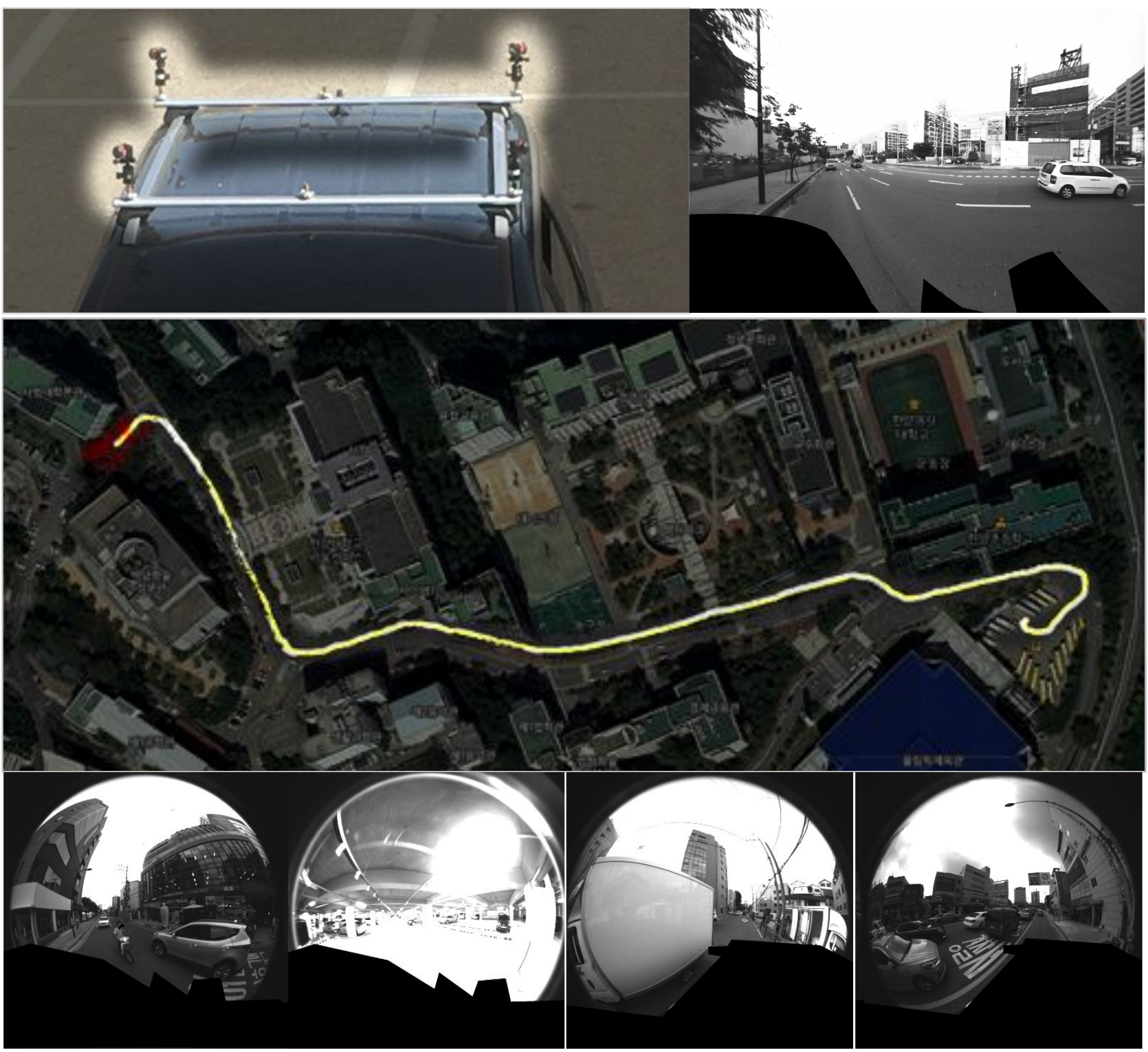}
\caption{Top: Our system setup. Four cameras are mounted to the rigid rig on the vehicle rooftop (left). The warped image with our hybrid projection model (right).
Middle: the estimated metric trajectory overlaid on the satellite image.
Bottom: four wide-FOV input images in challenging real datasets.
}
\label{fig:figure_intro}
\vspace{-5ex}
\end{figure}

In this paper we propose a novel omnidirectional visual odometry system using a multi-camera rig with wide FOV fisheye lenses and wide baseline, to maximize the stability and accuracy of the pose estimation.
From four cameras with 220\degree\ FOV lenses, it is possible to observe full 360\degree\ angle around the robot, and most regions in the environment are visible from more than two views which makes the stereo triangulation possible.

Our wide-baseline wide-FOV setup maximizes the pose accuracy as the stereo resolution is proportional to the baseline length, and it also minimizes the image area occluded by the robot body.
However it also poses new challenges.
First tracking and matching features become harder because the viewing directions to the points can be substantially different, and there is large distortion in the periphery of fisheye images where most overlaps between views occur.
Second, as the cameras are mounted far apart, the rigid rig assumption is no longer valid. 
Due to shocks, vibrations, and heat the rig can deform or the cameras can move unexpectedly.

We add three novel components to the VO pipeline to resolve these issues.
To improve the feature matching the lens distortion needs to be removed, but as the fisheye lens covers more than 180\degree, the image cannot be warped to a single plane.
We propose a hybrid projection model that uses two planes in the overlapping regions and a cylinder smoothly connecting them.
This projection model enables continuous tracking of feature points in each view and consistent feature descriptors across views. 

To estimate the current pose from noisy feature matches, we need to use a RANSAC algorithm with a minimal pose estimator.
For the omnidirectional case, one can use minimal solvers with a generalized camera model~\cite{Kneip2014UPnPAO,Schweighofer2008GloballyOO}, but we use a simpler approach of computing the pose using P3P~\cite{kneip2011novel} from one view and checking the inliers for all feature matches.
This multi-view P3P RANSAC effectively and robustly estimates the rig poses in highly dynamic scenes.

Lastly we implement the online extrinsic calibration to deal with unexpected changes of rig-to-camera poses throughout the system execution.
Besides the rig deformation by external causes, the initial calibration may not be very accurate due to the size and position of the rig.
Online extrinsic calibration built in the local bundle adjustment constantly updates the extrinsic parameters from the tracked features in the current map, and it greatly improves the robustness and accuracy of the system.

For experimental evaluation we render synthetic datasets with ground-truth poses as well as collect challenging real datasets using the omnidirectional rig in Figure~\ref{fig:figure_intro}.
Extensive experiments show that the proposed VO system accomplish good performance in all synthetic and real datasets.

Our main contributions can be summarized as follows:
\begin{itemize}
\item An effective novel image projection model which allows to find and track feature correspondences between the fisheye cameras in a wide-baseline setup.
\item A proposed visual odometry system that uses multiple fisheye cameras with overlapping views operates robustly in highly-dynamic environment using the multi-view P3P RANSAC algorithm and the online extrinsic calibration integrated with the backend local bundle adjustment. 
\item Extensive evaluation of the proposed algorithms to the synthetic and real datasets verifies the superior performance. All datasets as well as the system implementation will be made public with the paper publication.
\end{itemize}

\section{RELATED WORKS}

In the VO and visual SLAM literature, many different camera configurations have been researched. 
There are various monocular systems~\cite{mur2015orb,engel2018direct,forster2014svo} that are point feature-based, directly optimizing poses with image contents, or hybrid. 
They show outstanding performance, but due to fundamental limitation of monocular setup, metric poses cannot be estimated. 
For robotic application, stereo-based systems~\cite{mur2017orb,wang2017stereo} have been proposed. 
Another limitation of the conventional systems is small FOV which can make a VO system unstable due to lack of features or existence of dynamic objects. 
For this practical reason, fisheye camera based methods have been researched recently. 
Caruso et al.~\cite{caruso2015large} propose a fisheye visual SLAM with direct methods. 
Liu et al.~\cite{liu2017direct} use a fisheye stereo camera and recover metric scale trajectory. 
Most recently, Matsuki et al.~\cite{matsuki2018omnidirectional} proposed an omnidirectional visual odometry with the direct sparse method. 

For improved environmental awareness and perception capabilities, multi-camera methods also have been studied. 
\cite{hee2013motion} present a visual odometry algorithm for a multi-camera system which can observe full surrounding view. 
They successfully estimate ego-motion with the 2-point algorithm showing the importance of the inter-camera correspondences to recover metric scale. 
Heng et al.~\cite{heng2015self} implement a visual SLAM and self-calibration system with at least one calibrated-stereo camera and an arbitrary number of monocular cameras where they have overlapping views with the stereo camera.  
Recently, a robust multi-camera system using direct methods with plane sweeping stereo is proposed by Liu et al.\cite{liutowards}. 
Finding correspondences between fisheye images is a challenging and important problem and many researchers devoted efforts in it. 
Special descriptors~\cite{guan2017brisks,zhao2015sphorb} are designed to consider the distortion, and Hane et al.~\cite{hane2014real} and Gao et al.~\cite{gao2017dual} proposed dense matching algorithms for fisheye images.

\begin{figure}[tb]
\centering
\includegraphics[width=0.66\linewidth]{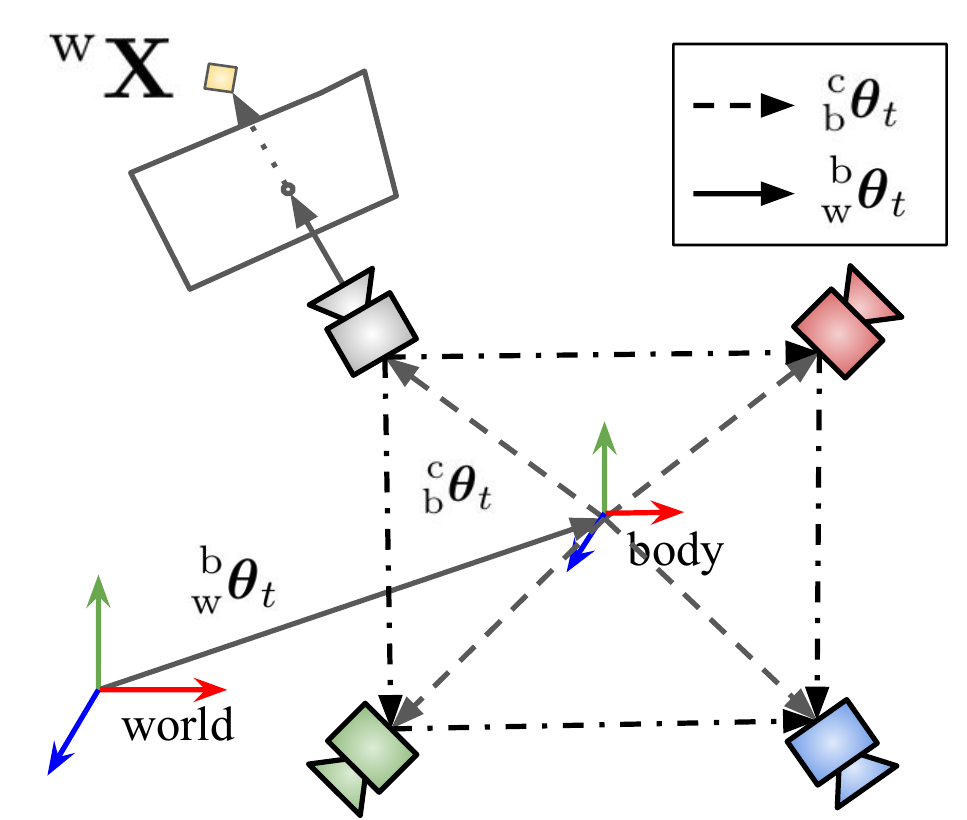}
\captionsetup{justification=centering}
\caption{The world, rig(body), and camera coordinate systems and transformations between them.}
\label{fig:system_rig}
\vspace{-3ex}
\end{figure}

\section{NOTATION}

In this section, we introduce the notations used in this paper. 
A rigid transformation $\btheta$ is parameterized as an axis-angle rotation vector $\br$ and a translation vector $\bt$ in $\mathbb{R}^3$.
It transforms a 3D point $\bX$ to $\btheta\star\bX = R(\br)\bX + \bt$, where $R(\br)$ is the $3\times 3$ rotation matrix for $\br$.
$\star$ also denotes the composition of transformations, and $^{-1}$ is the inverse transformation.
As in Figure~\ref{fig:system_rig} we use three coordinate systems, world (w), body (b), and camera (c), and when needed the coordinate system is marked on the left of symbols, like $\csT{\rw}{\rb}{\btheta}$ meaning the rigid transform from the body to the world coordinate system or $\csw{\bX}$ a point in the world coordinate system.
When the time is involved it is denoted as a subscript.
For example the camera coordinate of a world point $\bX$ at time $t$ can be written as 
\[ \cscam{\bX}_t = \csT{\rc}{\rb}{\btheta}_t \star \csT{\rb}{\rw}{\btheta}_t \star \csw{\bX}. \]

The camera intrinsic calibration determine the mapping between a 3D point $\cscam{\bX}$ and a pixel coordinate $\bx$ in the image.
We denote the projection function $\bx = \pi(\cscam{\bX};\bphi)$ for a camera intrinsic parameter $\bphi$. 
We use $\pi_0(\cdot)$ for projection onto the unit sphere, $\bar{\bx} = \pi_0(\bX)$, where $\bar{\bx}$ is a unit-length ray pointing $\bX$, which can also be a feature point in the image.
\begin{figure}[t]
\centering
\vspace{3ex}
\includegraphics[width=\linewidth]{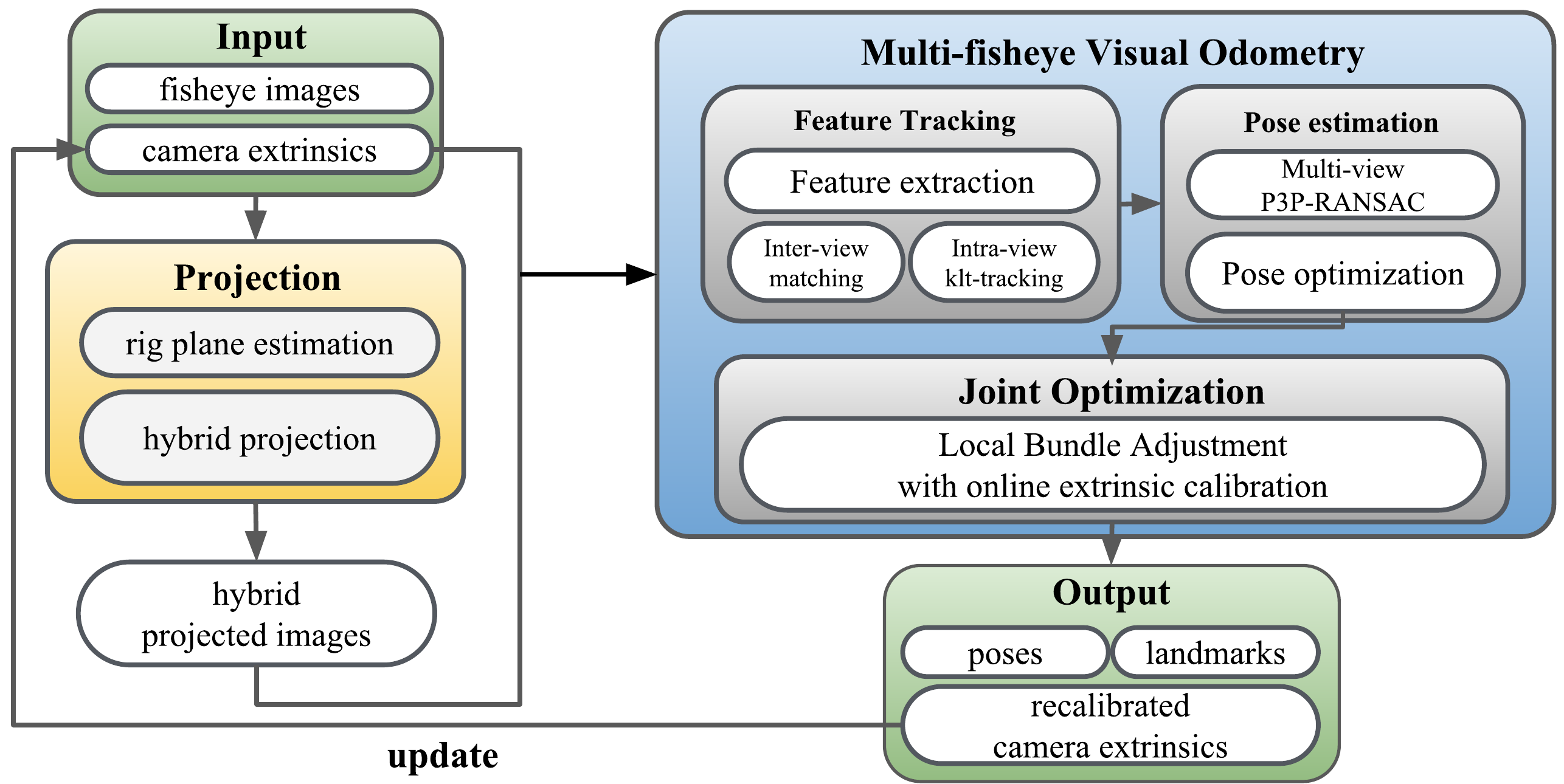}
\caption{Overview of the proposed system. 
The input fisheye images are warped using the hybrid projection model, then the multi-camera VO module tracks and matches the features, computes the current pose, and optimizes the map and the extrinsics. See Section~\ref{sec:algorithm} for detailed description.
}
\label{fig:overview}
\vspace{-4ex}
\end{figure}

\section{ALGORITHM}
\label{sec:algorithm}

We propose the robust omnidirectional visual odometry (ROVO) system with hybrid projection model, multi-view P3P RANSAC, and online-extrinsic calibration. 
An overview of the system architecture is show in Figure~\ref{fig:overview}. 
We assume that the camera intrinsic parameters and the initial extrinsic parameters are known.
When input fisheye images arrive, the hybrid projection model warps the input images into perspective-cylindrical images, and the feature detection and tracking modules run on the warped images per view.
After the intra-view feature tracking, we perform inter-view feature matching by comparing the feature descriptors in the overlapping regions. 
Then the multi-view P3P RANSAC algorithm computes the real-scale camera pose, followed by pose optimization which updates the estimated rigid pose with the 3D points fixed.
Finally the back-end local bundle adjustment module updates the recent rig trajectory and 3D point locations, as well as the rig extrinsic parameters in a unified framework.
For optimization tasks we use Ceres solver~\cite{ceres-solver}.

\subsection{Hybrid Projection Model} 
\label{sec:hybrid_projection_model}

\begin{figure}[t]
\centering
\begin{subfigure}{0.45\textwidth}
\includegraphics[width=\linewidth]{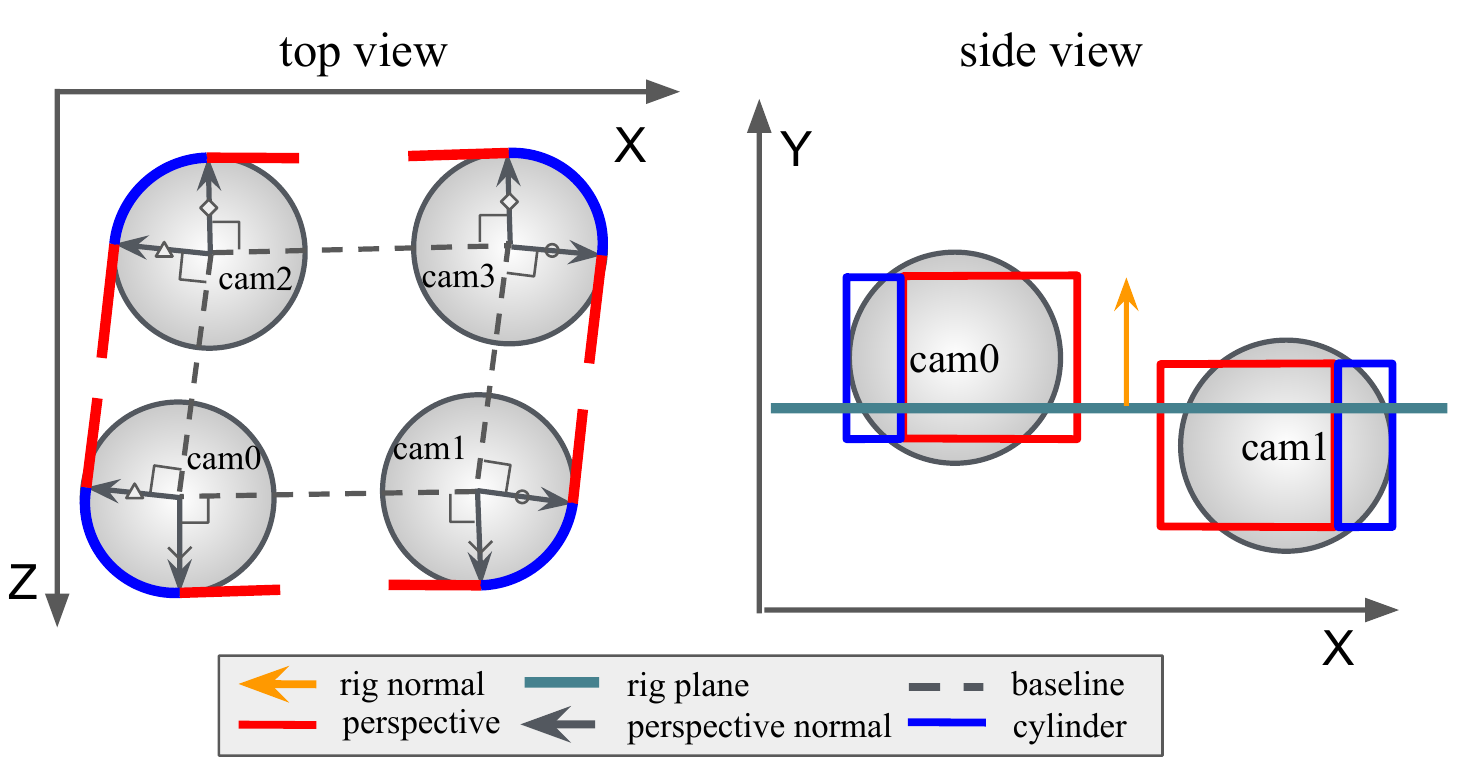}
\caption{Illustration of the hybrid projection model. The rig plane is best fitting plane given relative position of cameras. }
\label{fig:hybrid_projection}        
\end{subfigure}
\begin{subfigure}{0.45\textwidth}
\includegraphics[width=\linewidth]{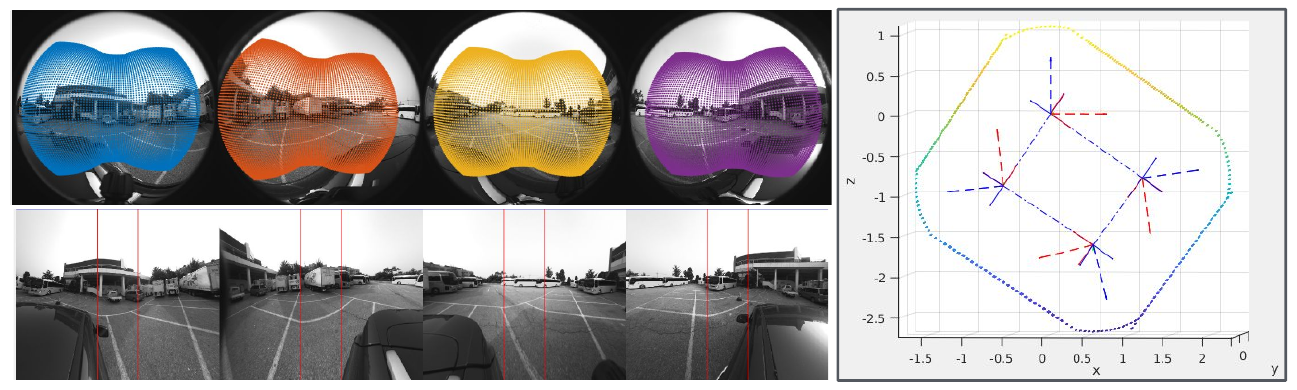}
\caption{
 Left: The pixels mapped in the original fisheye images (top) and the warped result image (bottom). Right: the top view of the projection planes and cylinders.
}
\label{fig:projection_fig}  
\end{subfigure}
\caption{Hybrid projection model. See Section~\ref{sec:hybrid_projection_model}}
\vspace{-4ex}
\end{figure}

Due to wide FOV the original fisheye images contain large amount of information about the environment, but at the same time the periphery of the images is distorted excessively.
In our wide-baseline wide-FOV setup there exist large overlapping area between images, but direct feature matching with the descriptors from such distorted areas does not yield good correspondences in quality and quantity.
The local feature descriptors work best when the images are purely perspective, thus we develop a projection model which ensures the overlapping areas between views are as perspective as possible.
At the same time, for feature tracking to be successful, the warped image must be continuous and smooth.

To satisfy these conditions, our hybrid projection model has two projection planes at the left and right sides and the cylinder at the center connects the two planes smoothly. 
Figure~\ref{fig:hybrid_projection} shows the hybrid projection model for our rig; for each camera the planes parallel to the baselines of the cameras are connected with cylinders perpendicular to the projection planes.
Note that when the camera centers are located close to the rig plane, the proposed method is similar to the stereo rectification and the projected $y$-coordinates of a scene point in the other images should be roughly same.

To build the warped image, we need to find the pixel position $\hat{\bx}$ in the original fisheye image corresponding to each pixel $\bx$ in the warped image.
For the point $\bx$, the 3D point $\cscam{\bPi}(\bx)$ on the projection plane/cylinder can be computed using the plane/cylinder equation, and then its fisheye image coordinate is given as
$ \hat{\bx} = \pi(\cscam{\bPi}(\bx); \bphi) $. \\
Figure~\ref{fig:projection_fig} shows the fisheye coordinates for the warped pixels (top) and the warped images (bottom).

\subsection{Intra- and Inter-view Feature Processing} 

For each warped image, we perform the intra-view feature processing which is the standard feature detection and tracking.
We use the ORB feature detector~\cite{rublee2011orb} with minimum distance constraints to neighboring features to ensure that features are extracted uniformly. 
The existing feature points are tracked by the KLT tracking algorithm~\cite{lucas1981iterative}. 

To improve the pose estimation and mapping quality we need to track the features across views.
The inter-view feature processing finds the matching features in the overlapping regions between views and transfer the information when a feature goes out of the FOV in one image.
The ORB descriptors are attached to the tracked points, and we use the $k$-nearest neighbor feature matching algorithm to find the feature correspondences. 
Incorrect matches are filtered according to the policies similar to the stereo matching:
\begin{itemize}
\item $y$-distance between the two matched points is small.  
\item the feature discriptor and orientation difference is small. 
\item it fulfills epipolar consistency, left-right consistency and positive disparity.
\item the zero normalized cross correlation (ZNCC) cost is small.
\end{itemize}
Finally, we triangulate the matched points to compute their 3D coordinates, which are used in pose estimation.
After the feature processing, we obtain the feature-landmark correspondences
$\{\{(\bar{\bx}_{i_j}, \csw{\bX}_{i_j})\}_{i_j}\}_j$ for all camera $j$'s. 
Note the feature locations are convertd to unit-length rays.

\subsection{Multi-view P3P RANSAC} 

After the feature processing and triangulation, the current pose of the rig is estimated from the established 2D-3D feature correspondences.
Our multi-view P3P RANSAC algorithm extends the monocular P3P RANSAC algorithm~\cite{lepetit2009epnp}.
In our RANSAC iterations, one view is selected randomly with Probability Proportional to the Size of match sets (PPS sampling), then the minimum sample set is randomly chosen among the correspondences in the view.
From the camera pose candidates estimated by the monocular P3P algorithm, the rig poses are computed and all correspondences in all views are tested for inlier check.
The detailed process is shown in Algorithm~\ref{alg:4viewp3p}.

PPS sampling choose the cameras with more feature matches more frequently to increase the chance of finding good poses, while all-view inter checking enforces the estimated pose is consistent with all observations.
To determine the best pose in the RANSAC loop, we use the reprojection errors of the inlier matches only.

After RANSAC finishes, the estimated rig pose $\csT{\rb}{\rw}{\btheta}$ is optimized by minimizing the reprojection error of all inliers while the 3D points are fixed,
\[
 \min_{\csT{\rb}{\rw}{\btheta}} \sum_j \sum_{i_j^*} \rho\left(|| \bar{\bx}_{i_j^*} - \pi_0(\csT{j}{\rb}{\btheta} \star \csT{\rb}{\rw}{\btheta} \star \csw{\bX}_{i_j^*}) ||^2\right),
\]
where $\csT{j}{\rb}{\btheta}$ is the transformation from the body to the camera $j$'s coordinate system, $i_j^*$'s are the inlier indices, and $\rho$ is the Cauchy loss function which minimizes the influence of outliers.

\begin{algorithm}
\SetAlgoLined
\KwData{2D feature locations and 3D landmark correspondences in all views, $\{\{(\bar{\bx}_{i_j}, \csw{\bX}_{i_j})\}_{i_j}\}_j$}
\KwResult{Rigid body transformation $\csT{\rb}{\rw}{\btheta}^*$}
 \While{$iter < iter_{max}$}{
  Select a camera $j'$ using PPS sampling.
  
  Randomly sample 3 pairs in the selected camera.
  
  Get camera pose candidates $\{\csT{j'}{\rw}{\btheta}_k\}$ by P3P.
  
  Compute the body pose candidates $\{\csT{\rb}{\rw}{\btheta}_k\}$.
  
  \For{each body pose candidate $\csT{\rb}{\rw}{\btheta}_k$}{
   \For{each camera $j$} {
    Compute the camera pose $\csT{j}{\rw}{\btheta}_k$.
    
    Add reprojection error of all inliers in $j$:\\
      \hspace{-3ex} $C_k \,+\!\!= \!\sum_{i_j}\!\max(0, \tau_r - || \bar{\bx}_{i_j} - \pi_{0}(\csT{j}{\rw}{\btheta}_k \star \csw{\bX}_{i_j}) ||)$
  }
  \If{$C_k$ is largest}{
    $\csT{\rb}{\rw}{\btheta}^* \gets \csT{\rb}{\rw}{\btheta}_k$.
    
    Update $iter_{max}$ with the new inlier ratio.
  }
 }
}
\caption{multi-view P3P RANSAC algorithm}
\label{alg:4viewp3p}
\end{algorithm}

\subsection{Online Extrinsic Calibration}

To deal with the deformation and motion of the camera during operation, the camera extrinsic parameters are jointly updated in the local bundle adjustment module.
For online extrinsic calibration, we add the camera extrinsic parameters $\{\csT{\rc}{\rb}{\btheta}\}_\rc$ into the optimization
\[
 \min
 \sum_t \sum_j \sum_{i_j} \omega_{i_j} \rho\left(|| \bar{\bx}_{i_j,t} - \pi_0(\csT{j}{\rb}{\btheta} \star \csT{\rb}{\rw}{\btheta}_t \star \csw{\bX}_{i_j}) ||^2\right), \vspace{-1ex}
\]
where the rig poses $\{\csT{\rb}{\rw}{\btheta}_t\}_t$ in the active time window,
the landmark positions $\{\csw{\bX}_i\}_i $, as well as the camera extrinsics $\{\csT{j}{\rb}{\btheta}\}_j$ are optimized to minimize the cost.
We give higher weight $\omega_{i_j}$ for the points observed in multiple cameras. 


Since the cameras in our system are fixed on a rig, we need to give an extra constraint that the distance between the cameras are constant.
Otherwise the metric scale reconstruction is not possible as the rig can grow or shrink over time.
The constraints can be written as $|| \csT{j}{i}{\bt} || = ||\csT{j}{i}{\bt}_0||$ for neighboring camera pairs $(i, j)$, where $\csT{j}{i}{\bt}$ represents the translation from the camera $i$ to $j$.

\begin{table*}[t]
\centering
\vspace*{+3ex}
\begin{tabular}{|c|c@{~~}c@{~~}c|c@{~~}c@{~~}c|c@{~~}c@{~~}c|c|c|}
\hline
\multirow{3}{*}{Sequences} & \multicolumn{3}{c|}{Translation RMSE(m)} & \multicolumn{3}{c|}{Average Inlier Ratio(\%)} & \multicolumn{3}{c|}{Average reprojection Error(\degree)} & \multirow{3}{*}{\# of frame} & \multirow{3}{*}{Length(m)} \\
 & NoisyExt & OnlineExt & GTExt
 & NoisyExt & OnlineExt & GTExt
 & NoisyExt & OnlineExt & GTExt & & \\ \hline
4view-MultiFOV & 30.01 & 0.60 & 0.23 & 28.1 & 47.8 & 51.0 & 0.60 & 0.19 & 0.13 & 2200 & 440 \\ \cline{1-1} \cline{11-12} 
Static Urban & 83.17 & 5.11 & 5.0 & 21.6 & 50.2 & 53.2 & 0.53& 0.37 & 0.25 & 3000 & 2000 \\ \cline{1-1} \cline{11-12} 
Dynamic Urban & 65.65 & 5.68 & 1.62 & 27.9 & 56.9 & 57.5 & 0.48 & 0.29 & 0.18 & 1000 & 1320 \\ \cline{1-1} \cline{11-12} 
Cloudy Urban & 42.28 & 1.43 & 0.42 & 31.1 & 59.1 & 60.7 & 0.51 & 0.31 & 0.22 & 300 & 350 \\ \cline{1-1} \cline{11-12} 
Sunset Urban & 25.86 & 1.92 & 0.37 & 36.4 & 51.8 & 59.8 & 0.59 & 0.23 & 0.21 & 300 & 350 \\ 
\hline
\end{tabular}
\caption{Quanitative evaluations with synthetic datasets. NoisyExt yields largely incorrect
trajectories while OnlineExt shows good performance close to GTExt.
See Section~\ref{sec:synthetic_experiments} for more details.}
\label{table:quantitative_result}
\vspace{-3ex}
\end{table*}

\begin{figure*}[t]
\vspace{-0.5ex}
\centering    
    \begin{subfigure}{0.25\textwidth}
    	\includegraphics[width=\linewidth]{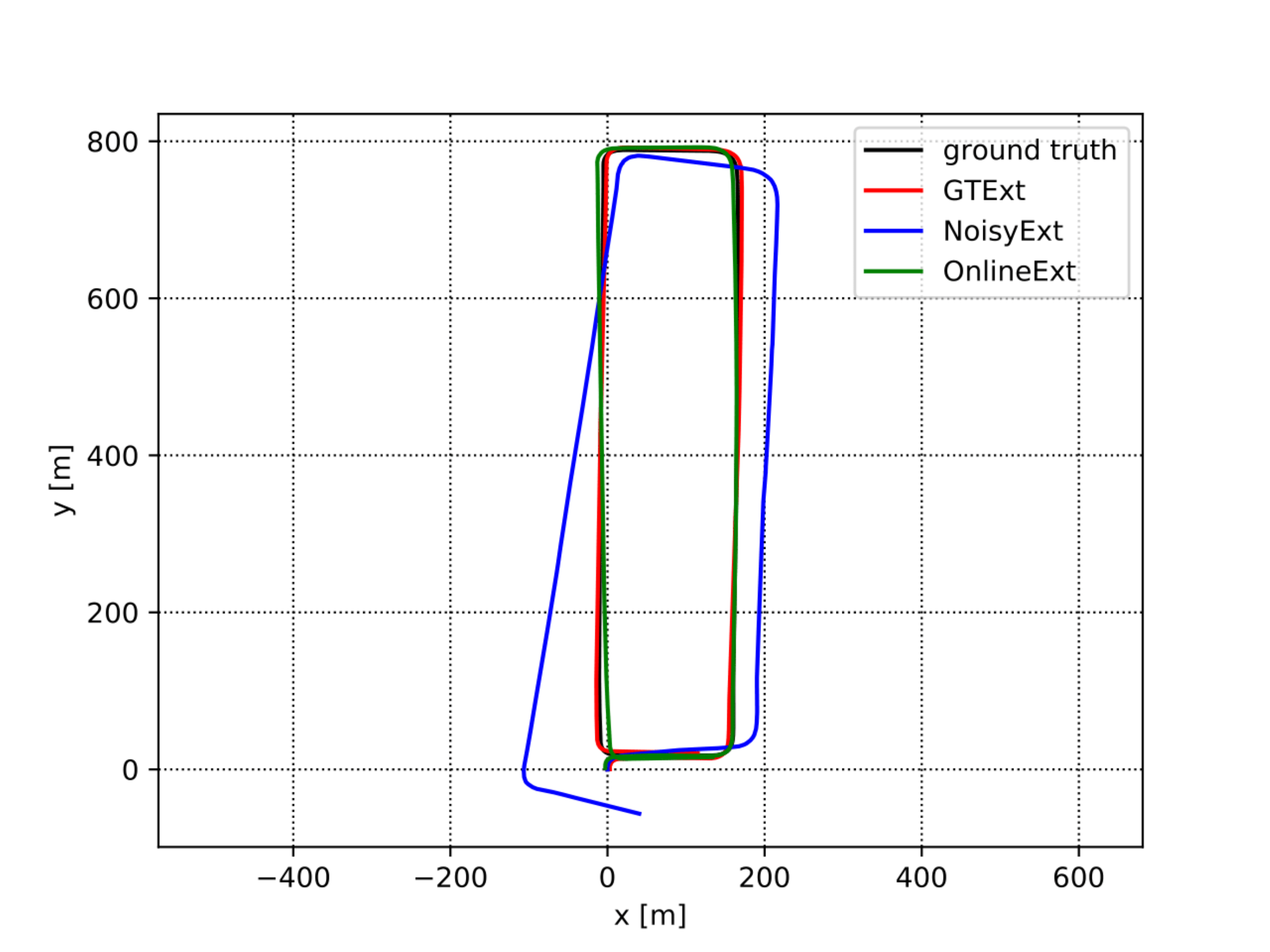}
        \captionsetup{justification=centering}
        \caption{Static Urban}
        \label{fig:result_static_urban}
    \end{subfigure}%
    \begin{subfigure}{0.25\textwidth}
    	\includegraphics[width=\linewidth]{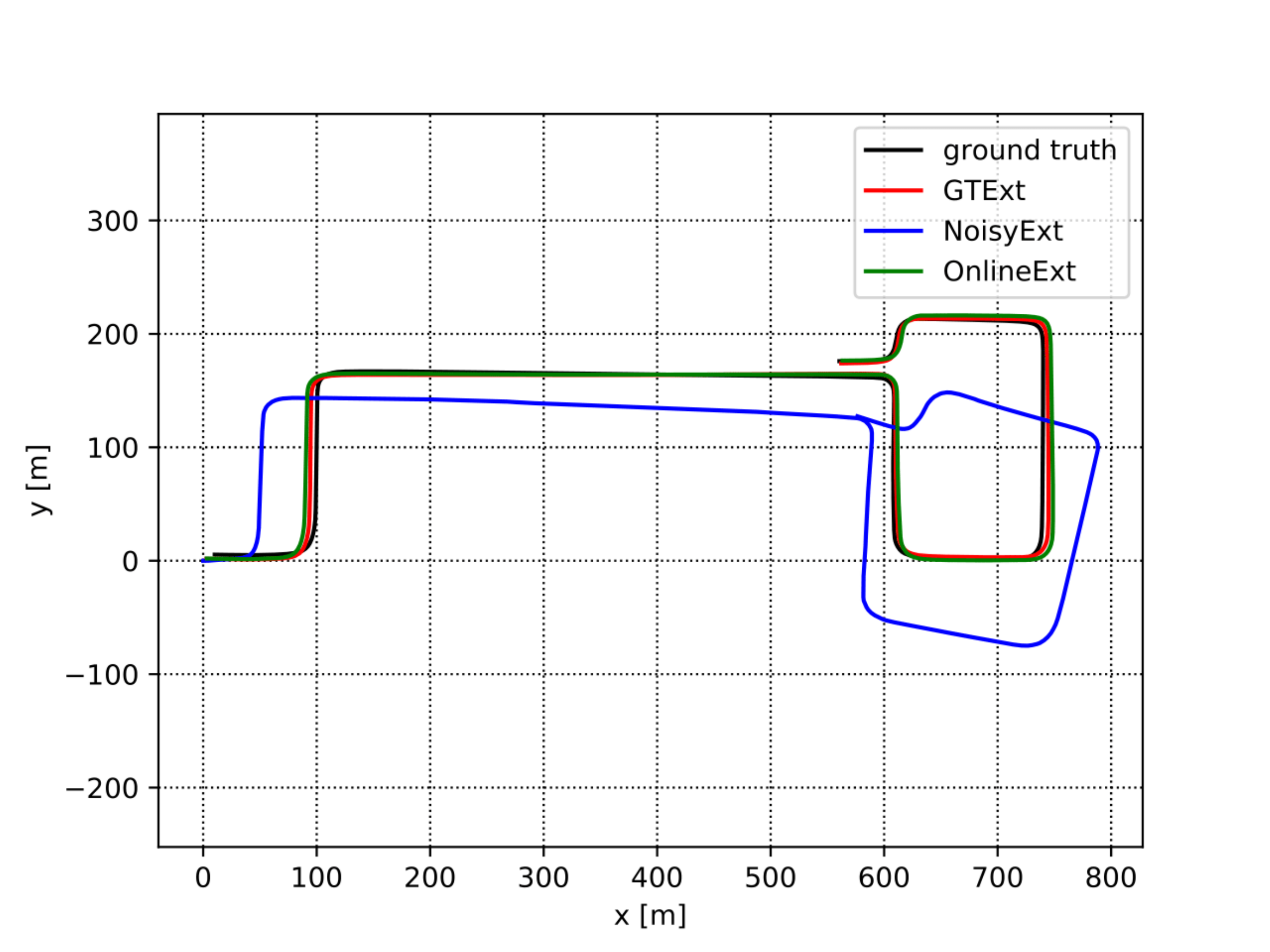}
        \captionsetup{justification=centering}
        \caption{Dynamic Urban}
        \label{fig:result_dynamic_urban}
    \end{subfigure}%
    \begin{subfigure}{0.25\textwidth}
    	\includegraphics[width=\linewidth]{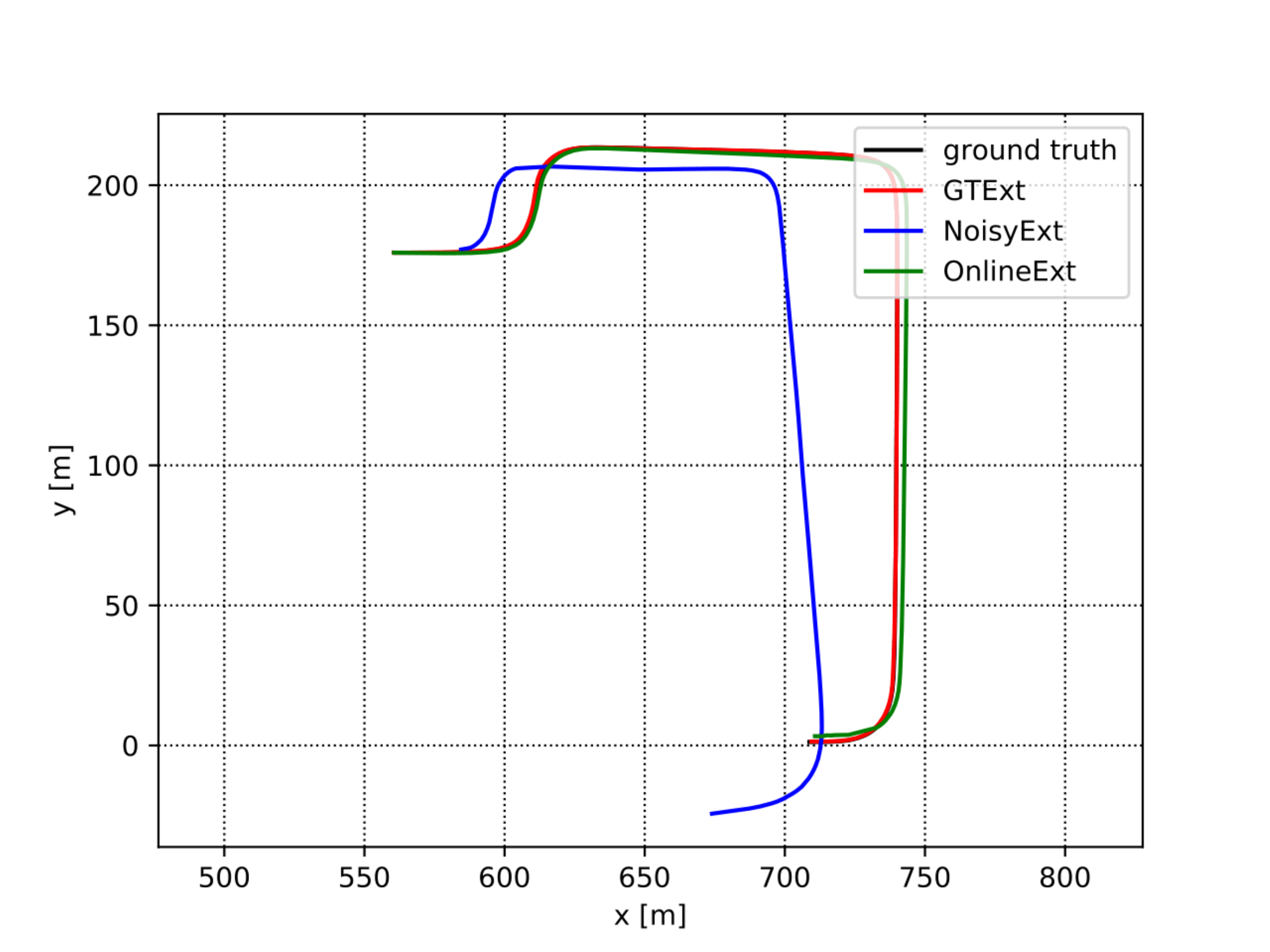}
        \captionsetup{justification=centering}
        \caption{Cloudy}
        \label{fig:result_cloudy}        
    \end{subfigure}%
    \begin{subfigure}{0.25\textwidth}
    	\includegraphics[width=\linewidth]{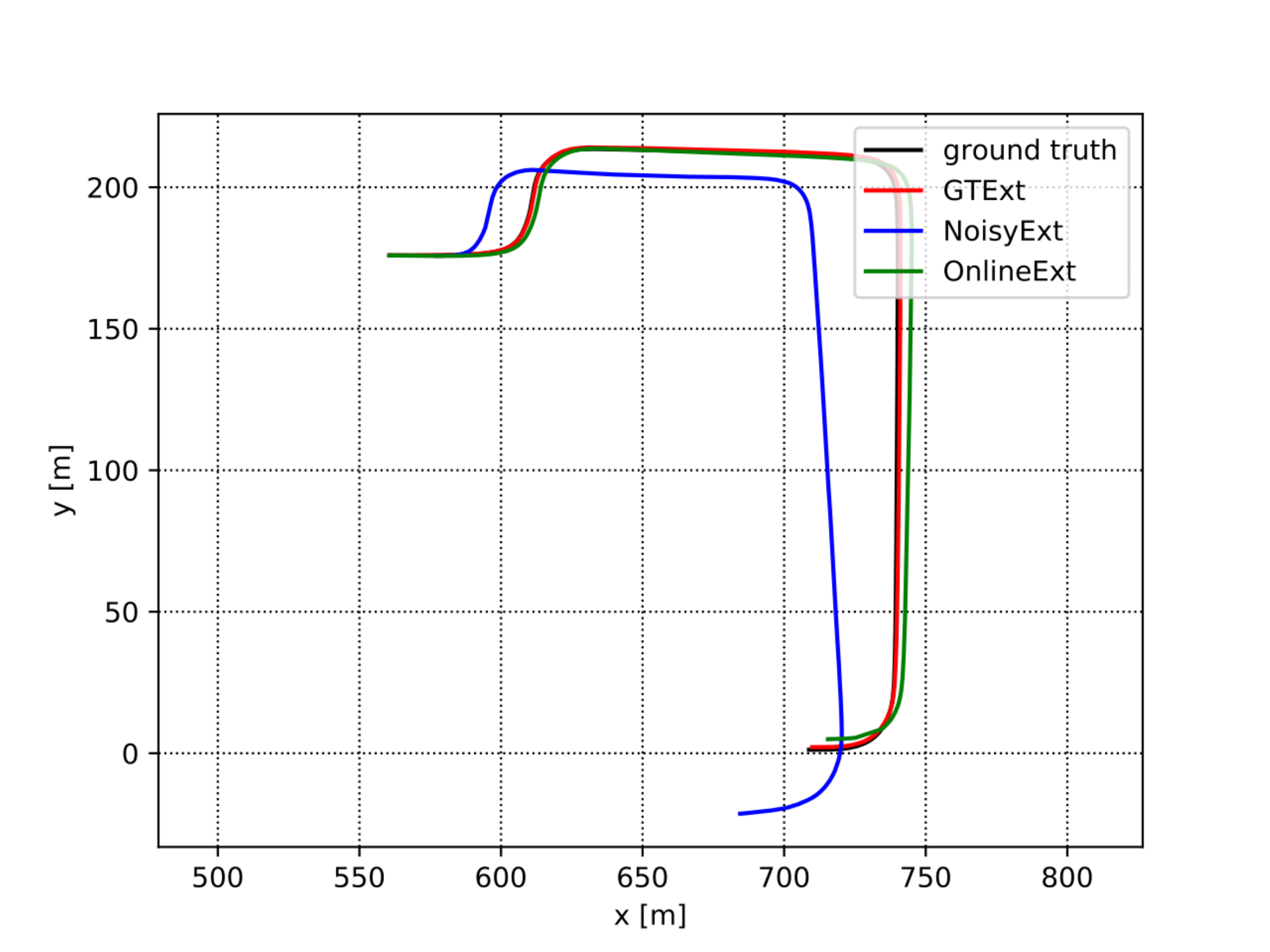}
        \captionsetup{justification=centering}
        \caption{Sunset}
        \label{fig:result_sunset}        
	\end{subfigure}
\caption{The estimated trajectories with the synthetic data-sets. The red, blue, green, and black lines represent the GTExt, OnlineExt, NoisyExt, and GT respectively. See Section~\ref{sec:synthetic_experiments} for more details.
}
\label{fig:synthetic_results}        
\vspace{-2ex}
\end{figure*}

\begin{figure}[t]
\centering
\vspace{-1.8ex}
\includegraphics[width=0.8\linewidth]{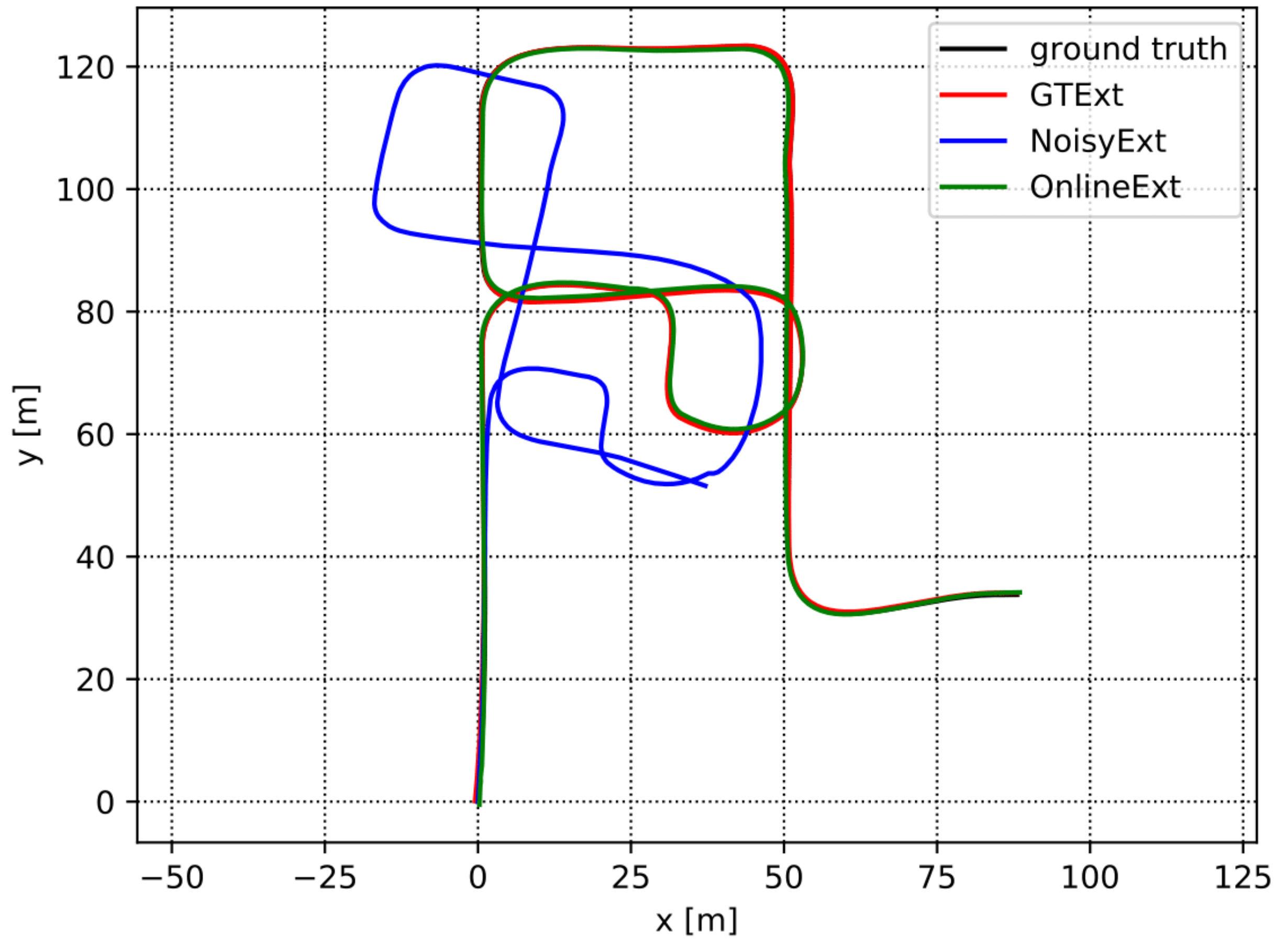}
\includegraphics[width=0.9\linewidth]{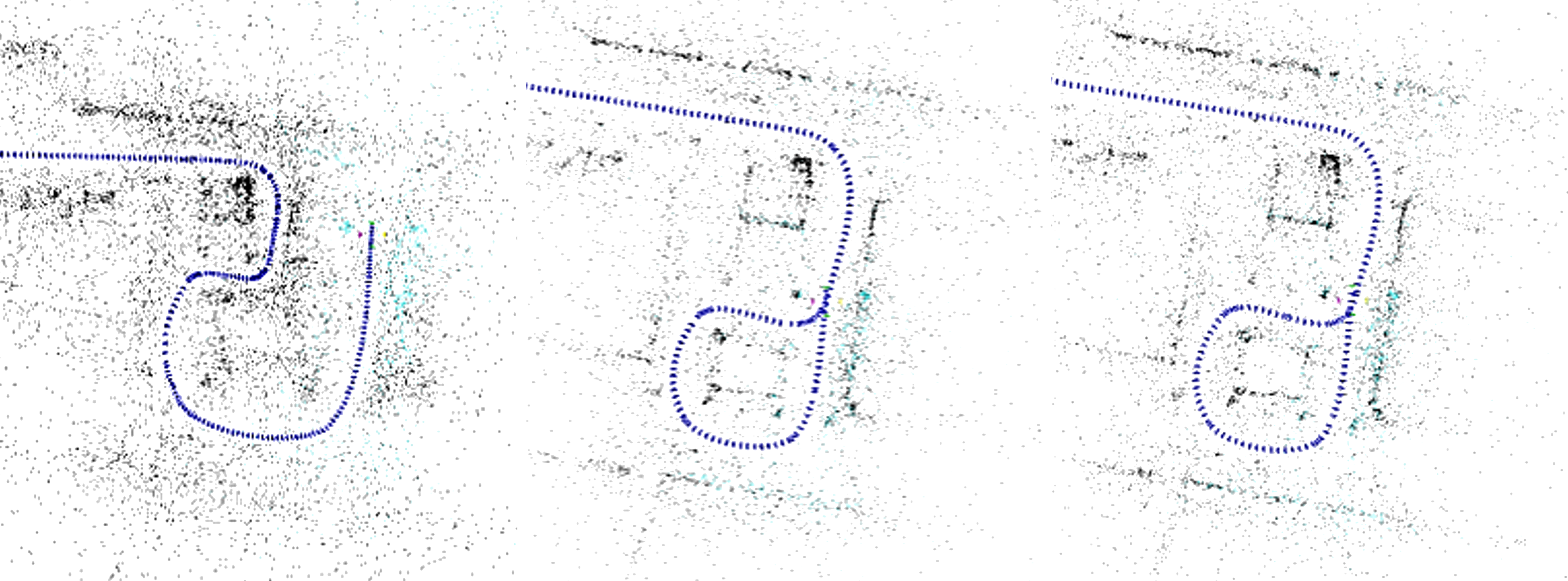}
\caption{
Experimental results in 4view-MultiFOV sequence. 
Top: the estimated trajectories. 
Bottom: NoisyExt (left), OnlineExt (center), GTExt (right).}
\label{fig:result_oncalib_qualitative}        
\vspace{-2ex}
\end{figure}

\begin{figure*}[t]
\centering
\vspace*{3ex}
\begin{subfigure}[t]{0.23\textwidth}
\includegraphics[width=\linewidth]{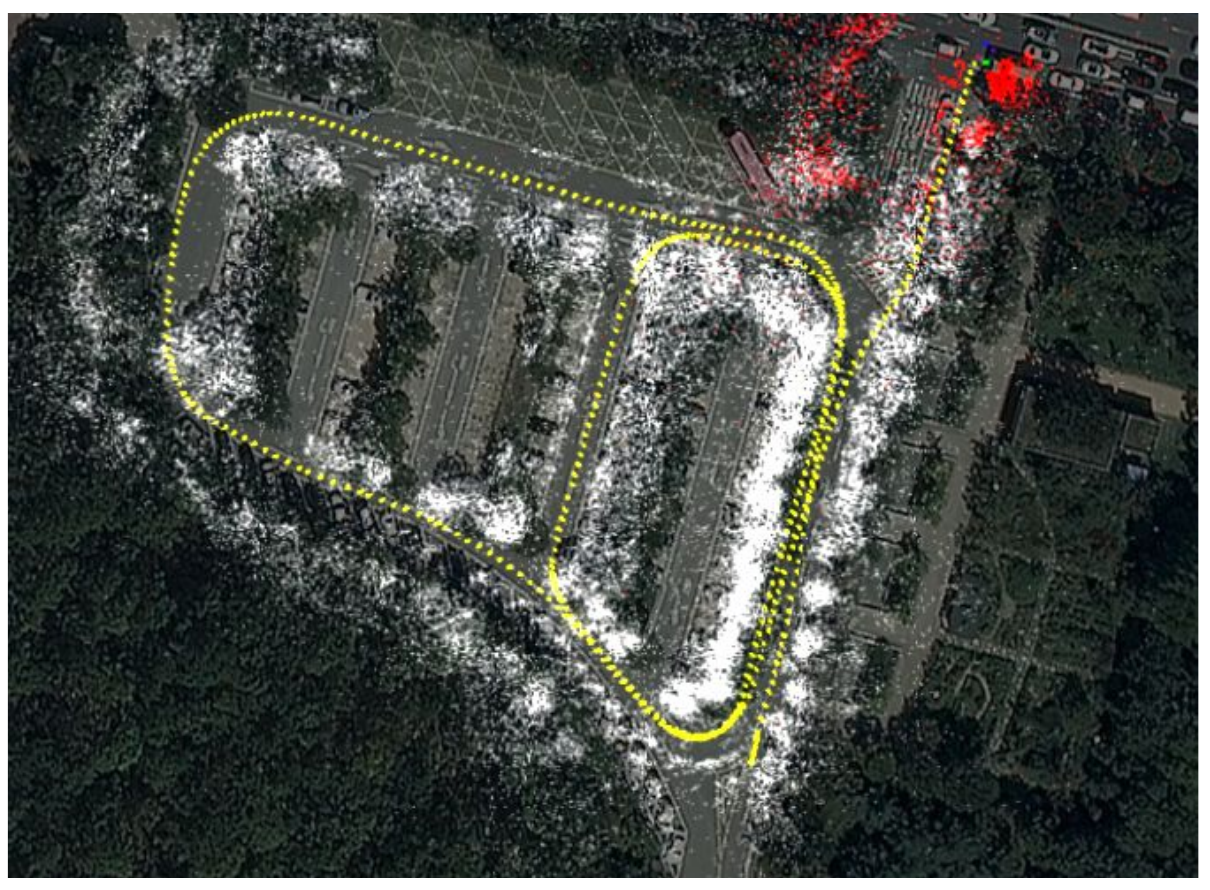}
\caption{ParkingLot sequence}
\end{subfigure}
\begin{subfigure}[t]{0.35\textwidth}
\includegraphics[width=\linewidth]{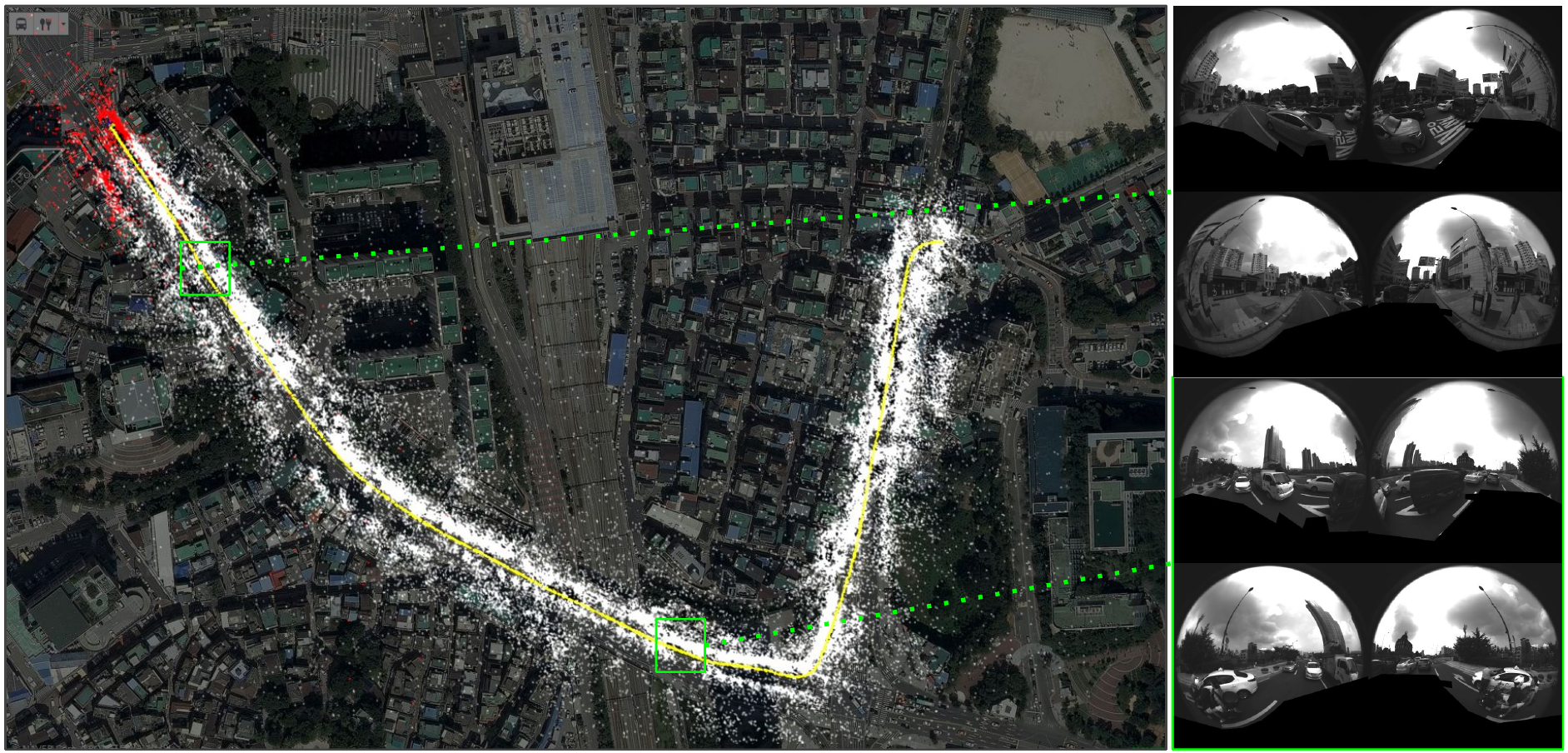}
\caption{Wangsimni sequence}
\end{subfigure}
\begin{subfigure}[t]{0.36\textwidth}
\includegraphics[width=\linewidth]{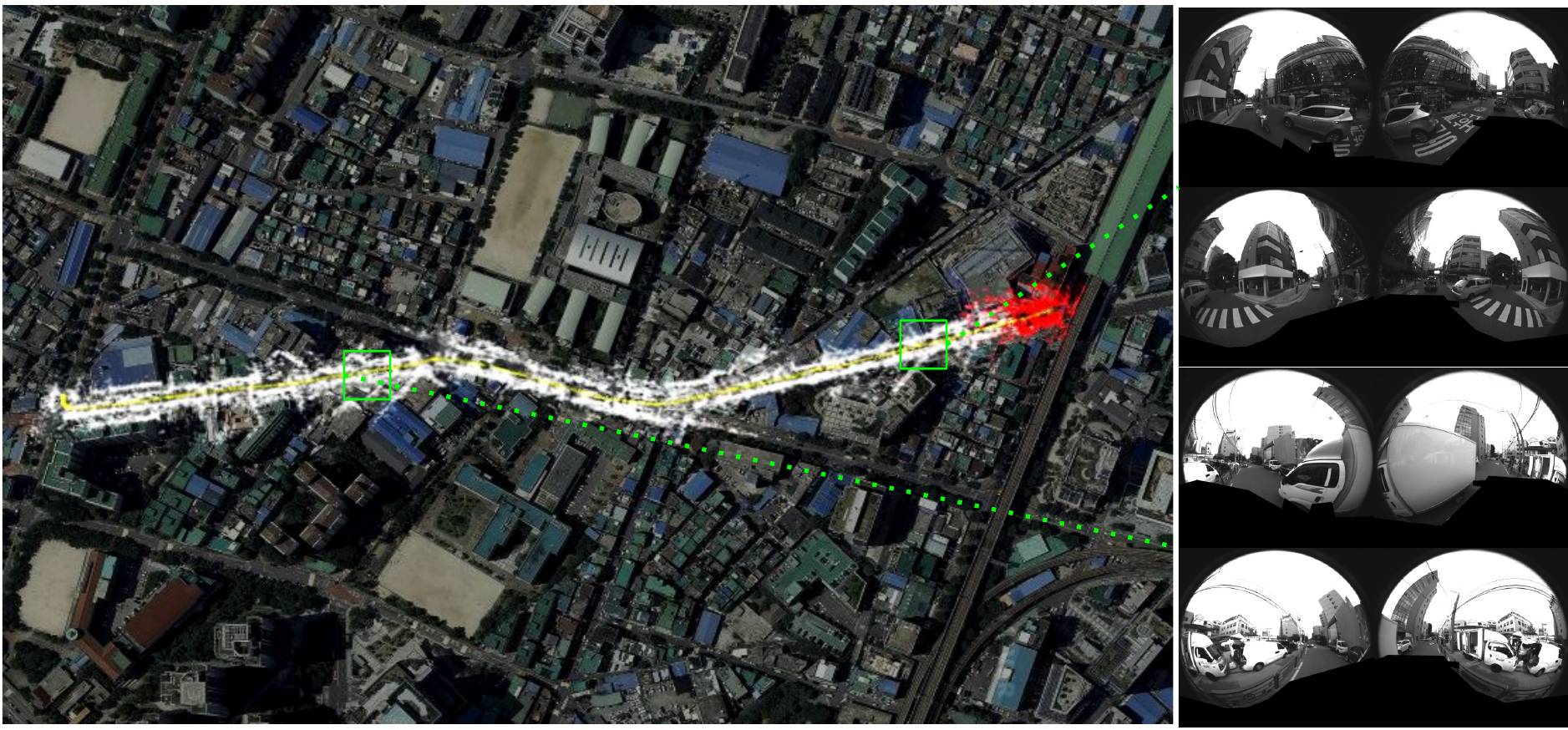}
\caption{Seongsu sequence}
\end{subfigure}
\caption{The estimated trajectories (yellow) of the real sequence overlaid on the satellite map. The white points represent the reconstructed 3D landmarks. See Section~\ref{sec:real_experiments} for detailed discussion.
}
\label{fig:real_results}        
\vspace{-3ex}
\end{figure*}

\begin{figure}[t]
\centering
\includegraphics[width=\linewidth]{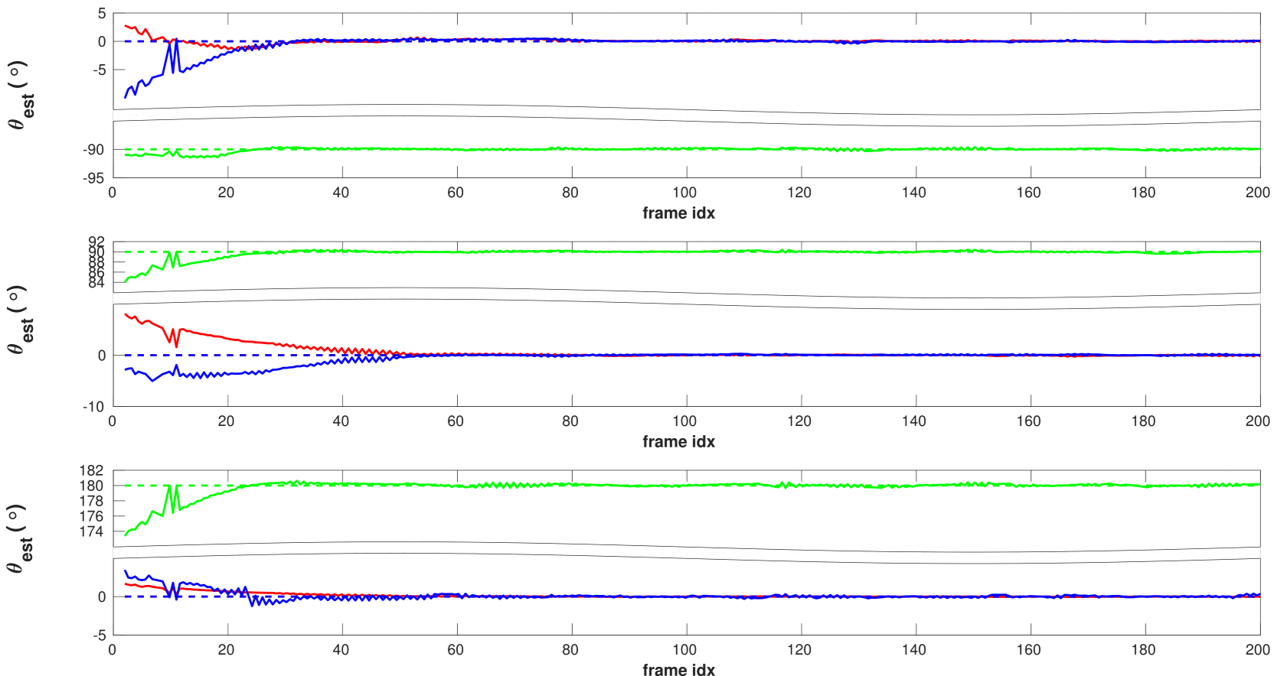}
\caption{
The three plots correspond to each camera's relative pose to the first camera.
Red, green, and blue lines represent pitch, roll, and yaw values respectively. The dotted line is the ground-truth. 
With online extrinsic calibration the initial noisy values quickly converges to the ground-truth.
}
\label{fig:result_oncalib_quantitative} 
\vspace{-3ex}
\end{figure}

\begin{figure}[t]
\centering
\vspace{-2ex}
\includegraphics[width=0.95\linewidth]{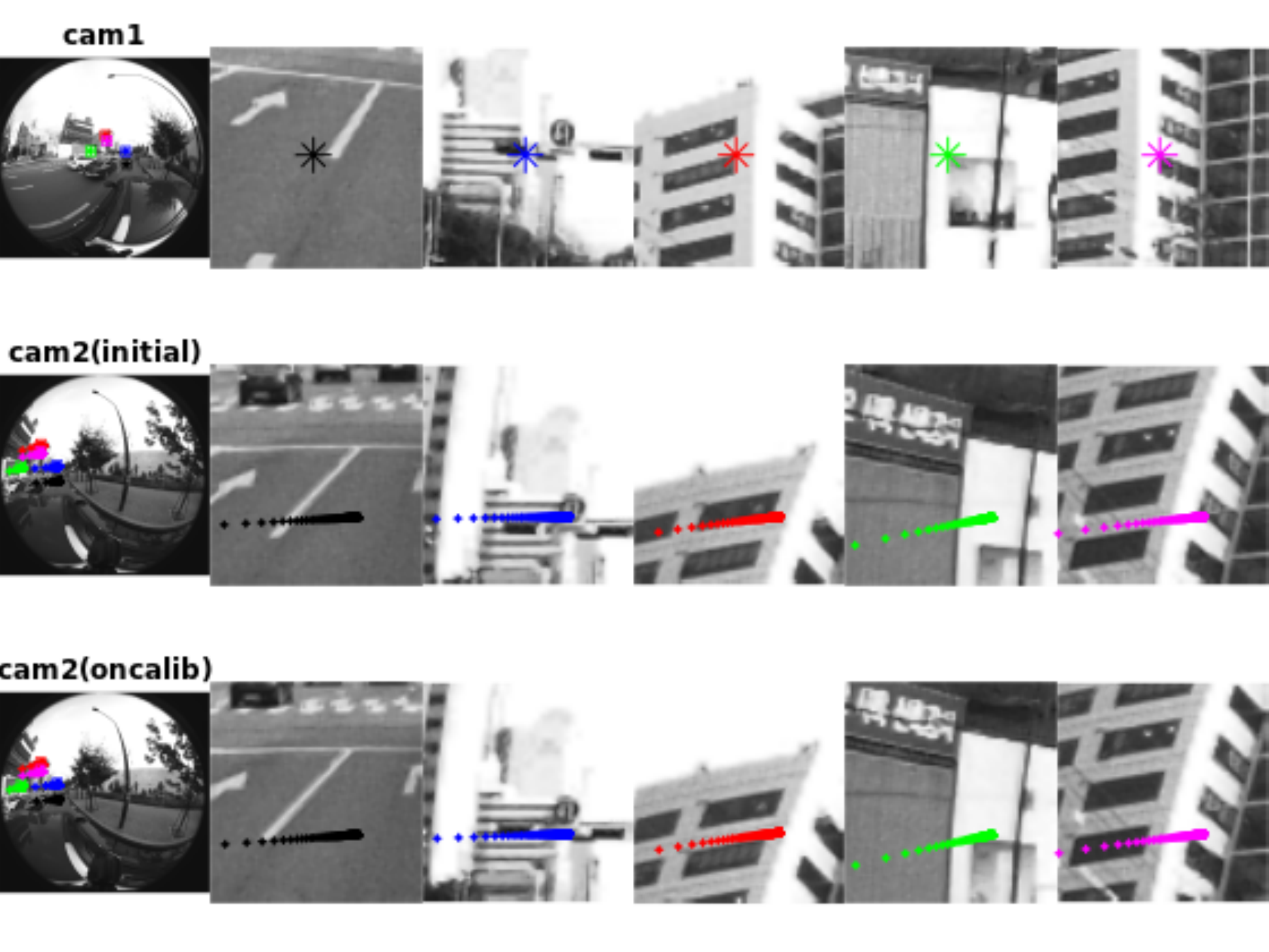}
\caption{
Online extrinsic calibration result in a real dataset. 
Top: sample point locations in the reference camera. 
Middle: epipolar lines by initial extrinsics in the target camera.
Bottom: corrected epipolar lines by online extrinsic calibration.
}
\captionsetup{justification=centering}
\label{fig:online_calib_real} 
\vspace{-2ex}
\end{figure}


\section{EXPERIMENTAL RESULTS}

 In order to evaluate the proposed system, we conduct extensive experiments with synthetic datasets along with real-world datasets. 
 Using the synthetic datasets with ground-truth, we quantitatively measure the accuracy by computing the Root Mean Squared Error(RMSE) between the estimated poses and the ground truth. 
 In addition, we compare average inlier ratio and average reprojection error to observe the tendency of experimental results. 
 In the real datasets, we qualitatively evaluate the performance by overlaying the estimated trajectory to the satellite images. 
 Additionally, we show the effectiveness of online extrinsic calibration in both synthetic and real datasets with comparative experiment.  

\subsection{Evaluation of Hybrid Projection Model}

To test our hybrid projection model, we conduct two simple experiments. 
First, we compare the ORB descriptor similarity during intra-view tracking.
The mean hamming distances of the 2000 features for 100 frames in the original fisheye images and in our hybrid-warped images are compared. 
Second, we compare the inlier ratios of feature correspondences between inter-view matching. 
The inlier ratio is computed only from the feature matches that satisfies the epipolar constraints. 

\begin{table}[t]
\vspace{-1ex}
\centering
\begin{tabular}{|c|c|c|}
\hline
 & Original fisheye & Hybrid projection (ours) \\ \hline
Descriptor distance & 124.85 & 35.27 \\ \hline
Inlier ratio(\%) & 21.53 & 73.16 \\ \hline
\end{tabular}
\captionsetup{justification=centering}
 \caption{Experiments of the hybrid projection model. 
}
\label{table:hybrid_projection_results}
\vspace{-5ex}
\end{table}

As shown in Table~\ref{table:hybrid_projection_results} our projection model reduces the average descriptor distance and boosts the inlier ratio more than $3\times$, mainly by removing the lens distortion and aligning the projections.
The boosted matching performance significantly contributes the robustness of VO systems.


\subsection{Synthetic and Real Datasets}

We render four urban sequences with different structures lighting conditions using Blender.
As a baseline we use 4view-MultiFOV which is modified from the urban canyon dataset by Zhang et al.~\cite{zhang2016benefit}.
Static Urban is a 2km-long sequence with a significant illumination change due to building shadows. 
Dynamic Urban is a 1.3km-long sequence with moving vehicles. 
Cloudy Urban and Sunset Urban are 350m long sequences of the same scene with different weather conditions. 
All images are rendered for four simulated fish-eye cameras ($220\degree$ FOV) with $1600\times1532$ resolution, which is same FOV and resolution with the real camera.
The ground truth poses, camera intrinsic parameters and extrinsic parameters are included in the datasets.

For real datasets we use the four global-shutter camera rig on the vehicle as shown in Figure~\ref{fig:figure_intro}. 
All cameras output four software-synchronized $1600\times1532$ images at 10Hz. 
We use a standard camera rig calibration with a large checkerboard.
The datasets are collected by  driving the vicinity of Hanyang university, and they contain many challenges of harsh illumination changes, highly dynamic road with many moving vehicles, and narrow streets. 

\subsection{Experiments with Synthetic Datasets}
\label{sec:synthetic_experiments}

To evaluate the robustness and accuracy, we conduct an experiment by providing a randomly perturbed camera extrinsics to the system (zero-mean Gaussian noise with $\sigma=5\degree$).
NoisyExt is the result with the noisy extrinsics, and OnlineExt is the result with online extrinsic calibration.
For comparison GT refers to the ground-truth rig trajectory, and GTExt is the VO result with the ground-truth extrinsics.

Quantitative and qualitative comparison is shown in Table~\ref{table:quantitative_result},  Figure~\ref{fig:synthetic_results} and Figure~\ref{fig:result_oncalib_qualitative}. 
While the VO with noisy extrinsics fails to estimate correct trajectories,
with online calibration overall error decreases drastically and the trajectory is also accurately estimated. 
We also observe that average inlier ratio and average reprojection error are significantly improved close to GTExt.
Figure~\ref{fig:result_oncalib_quantitative} gives in-depth view of online extrinsic calibration. 
Starting with noisy extrinsics, the extrinsic parameters are updated quickly and converges to the ground-truth within 100 frames.

\subsection{Experiments with Real Datasets}
\label{sec:real_experiments}

Among the collected datasets, we present the results of ParkingLot, Wangsimni, and Seongsu.
ParkingLot has a loop trajectory and Figure~\ref{fig:real_results}(a) shows the accuracy of our system (note that our system do not use loop closing).
Figure~\ref{fig:real_results}(b), (c) shows the Wangsimni and Seongsu results.
Wangsimni sequence is taken in a heavy traffic with many moving objects.
Our system is able to reject those outliers and robustly estimate the accurate trajectory.
In Seongsu sequence the vehicle is driven in narrow passages, and the distance to the near buildings is very close compared to the camera baseline.
The hybrid warping can successfully find the matching features in this challenging situations.

The online calibration makes a big contribution in real sequences.
Figure~\ref{fig:online_calib_real} shows the reprojected epipolar lines of a feature in a neighbor view.
With the initial calibration, the epipolar lines are quite off from the true matches, but with online calibration they fall on the correct position.


\section{CONCLUSION}
In this paper we propose a novel omnidirectional visual odometry system for a wide-baseline camera rig with wide-FOV lenses.
To deal with the challenges from the fisheye distortion and appearance changes due to wide-baseline, we add a hybrid projection model, a multi-view P3P RANSAC algorithm, and online extrinsic calibration in local bundle adjustment.
The extensive experimental evaluation using both synthetic datasets with ground-truth and real sequences verifies that the proposed components are effective in solving the problems.
\vspace{-1.8ex}

\section*{ACKNOWLEDGMENT}
\begin{small}
This research was supported by Samsung Research Funding \& Incubation Center for Future Technology under Project Number SRFC-TC1603-05, Next-Generation Information Computing Development Program through National Research Foundation of Korea(NRF) funded by the Ministry of Science, ICT(NRF-2017M3C4A7069369), and the NRF grant funded by the Korea government(MISP)(NRF-2017R1A2B4011928).
\end{small}


\bibliographystyle{IEEEtran}
\bibliography{IEEEabrv,IEEEexample}

\end{document}